\documentclass[conference]{IEEEtran}
\IEEEoverridecommandlockouts

\usepackage{url,hyperref,lineno,microtype,subcaption}
\usepackage{mathtools}
\usepackage{algorithm}
\usepackage{algorithmic}

\usepackage{xcolor}
\usepackage{graphicx}
\usepackage{booktabs}
\usepackage{amssymb,amsmath,amsthm}

\newcommand{\tool}{{\sc LLMchecker}}
\newcommand{\alphak}{$\alpha$-$k$-bounded}

\usepackage{tikz}
\usetikzlibrary{decorations.text, arrows.meta, positioning}
\usetikzlibrary{decorations.pathreplacing,calc,shapes,positioning}
\usetikzlibrary{shapes,shadows,arrows,positioning,angles,quotes}
\tikzset{My Arrow Style/.style={single arrow, fill=black!15, anchor=base, align=center,text width=2.3cm}}
\tikzstyle{arrow} = [thick,->,>=stealth]
\usetikzlibrary{calc,trees,positioning,arrows,fit,shapes,calc}

\tikzstyle{startstop} = [rectangle, rounded corners, minimum width=1.5cm, minimum height=0.5cm,text centered, draw=black, fill=red!30]
\tikzstyle{io} = [trapezium, trapezium left angle=70, trapezium right angle=110, minimum width=0.5cm, minimum height=0.5cm, text centered, draw=black, fill=blue!30]

\tikzstyle{process} = [rectangle, minimum width=3cm, minimum height=0.5cm, text centered, draw=black, fill=orange!30]
\tikzstyle{decision} = [diamond, minimum width=0.5cm, minimum height=0.1cm, text centered, draw=black, fill=green!30]
\tikzstyle{process2} = [rectangle, minimum width=1cm, minimum height=0.5cm, text centered, draw=black, fill=orange!30]
\tikzstyle{arrow} = [thick,->,>=stealth]

\usetikzlibrary{automata, positioning}
\usetikzlibrary{shapes.geometric, arrows, positioning}

\tikzstyle{block} = [rectangle, draw, text centered, rounded corners, minimum height=2em]
\tikzstyle{line} = [draw, -stealth, thick]
\tikzstyle{cloud} = [ellipse, draw, text centered, minimum height=2em, thick]
\tikzstyle{dashedcloud} = [ellipse, draw, dashed, text centered, minimum height=2em, thick]
\usetikzlibrary{positioning, arrows.meta, shapes.geometric, decorations.pathreplacing}

\AtBeginDocument{%
  }

\theoremstyle{definition}
\newtheorem{definition}{Definition}[section]
\begin{document}

\title{Bounded PCTL Model Checking of Large Language Model Outputs%
\thanks{This work is funded by the European Union under grant agreement number 101091783 (MARS Project).}}

\author{\IEEEauthorblockN{Dennis Gross}
\IEEEauthorblockA{\textit{Simula Research Laboratory}\\ Oslo, Norway \\
dennis@simula.no}
\and
\IEEEauthorblockN{Helge Spieker}
\IEEEauthorblockA{\textit{Simula Research Laboratory}\\ Oslo, Norway \\
helge@simula.no}
\and
\IEEEauthorblockN{Arnaud Gotlieb}
\IEEEauthorblockA{\textit{Simula Research Laboratory}\\ Oslo, Norway\\
arnaud@simula.no}}

\maketitle

\begin{abstract}
In this paper, we introduce \tool, a model-checking-based verification method to verify the probabilistic computation tree logic (PCTL) properties of an LLM text generation process.
We empirically show that only a limited number of tokens are typically chosen during text generation, which are not always the same.
This insight drives the creation of \alphak\ text generation, narrowing the focus to
the $\alpha$ maximal cumulative probability on the top-$k$ tokens at every step of the text generation process.
Our verification method considers an initial string and the subsequent 
top-$k$ tokens while accommodating diverse text quantification methods, such as evaluating text quality and biases.
The threshold $\alpha$ further reduces the selected tokens, only choosing those that exceed or meet it in cumulative probability.
\tool\ then allows us to formally verify the PCTL properties of \alphak\ LLMs.
We demonstrate the applicability of our method in several LLMs, including Llama, Gemma, Mistral, Genstruct, and BERT.
To our knowledge, this is the first time PCTL-based model checking has been used to check the consistency of the LLM text generation process.
\end{abstract}


\section{Introduction}
\emph{Large language models (LLMs)} have recently advanced across many fields~\cite{saxena2024automated,DBLP:journals/npjdm/GuevaraCTCFKMQGHACSMB24}, transforming critical domains like automated transportation~\cite{nie2025} and healthcare~\cite{khalid2024}. Trained on vast text corpora, LLMs generate human-like text by predicting tokens sequentially~\cite{DBLP:conf/nips/VaswaniSPUJGKP17}.

Unfortunately, LLMs do not have guarantees to avoid faulty behaviors~\cite{DBLP:journals/corr/abs-2305-11391,prompt4perf} such as copyright violations~\cite{DBLP:conf/emnlp/KaramolegkouLZS23}, human-being-related biases~\cite{ratz2023measuring}, providing wrong information~\cite{DBLP:conf/cikm/ChenFYWFL0LX23}, or creating text with low quality~\cite{chen2023exploring}.
Automated text quantification methods can quantify these different issues.
There exists work already that empirically test LLMs~\cite{DBLP:journals/corr/abs-2307-03109,DBLP:conf/kbse/XiaoXDJZ23,DBLP:journals/corr/abs-2402-14480,DBLP:journals/corr/abs-2402-13518}.

\emph{Bounded model-checking} is a formal verification technique that uses logical reasoning to verify the satisfaction of a given specification property on a model, by exploring a finite number of system state transitions~\cite{biere2021bounded}.
As example model, \emph{Discrete-Time Markov Chain (DTMC)} can be used to represent systems whose states change probabilistically in discrete time steps.
Naively modeling the text generation of LLMs for a given text and text length as a DTMC is theoretically possible, but it is highly inefficient.
The vast number of potential tokens at each time step~\cite{DBLP:journals/corr/abs-2307-09288} leads rapidly to a combinatorial explosion of possible successor text strings, each modeled as a DTMC state, making it impractical.
Furthermore, in evaluating text generation systems, \emph{probabilistic computation tree logic (PCTL)}~\cite{DBLP:journals/fac/HanssonJ94}, commonly utilized to query system properties in model checking, is not tailored to articulate properties regarding textual output.

In this paper, we first demonstrate that LLMs typically favor by default a limited selection of tokens as likely choices at each step of their text generation, with the vast majority of other tokens from a finite set rarely being considered.
Next, we demonstrate how to apply PCTL model checking to partially verify the LLM's text generation process by modeling it first into a formal model that PCTL is designed to handle, followed by its verification.

To model the text generation process, we focus on a given text input, limit the LLM's output to a maximal text length and the most likely tokens, and model each possible text string during this text generation as a state of a DTMC.
To allow formalization of PCTL specifications, we quantify the actual text strings in the states using user-specified text quantification methods, such as text similarity metrics~\cite{DBLP:journals/information/WangD20}, bias counting~\cite{DBLP:conf/www/Babaeianjelodar20}, text quality scores~\cite{DBLP:journals/information/EleyanOE20}, and sentiment metrics~\cite{DBLP:conf/fit/ShaikhDYS23}.

During the modeling process, we select the next tokens in each possible text generation state until their cumulative likelihood (given by the LLM's output distribution) exceeds a threshold $\alpha$ or we have already selected a maximum number of $k$ tokens.
Basically, we build the highest probability combination of output tokens in a best-first manner~\cite{DBLP:conf/aaai/0002B24} with bounded cumulative probability resp. number of tokens.
Then, we visit the new states and do the same until we generate text strings with the specified length.
Since the transition probabilities between states must sum to one, we create additional terminal states for the not-selected tokens at each step of the text generation process to reduce the total number of states by aggregating the probabilities of all remaining options.
We then build the corresponding transition from the current to the terminal state.
This approach, implemented in a tool called \tool\ , allows us to check, for instance, the reachability probability of text similarity (for copyright violation checking) in the $\alpha$-$k$-bounded text generation process of the LLM.

Our experiments realized with \tool\ show that it is possible to verify the $\alpha$-$k$-bounded text generation process of an LLM for PCTL properties based on a given text input and maximal text length. %
Furthermore, we show \emph{exactly} how the properties of various LLMs, including Llama, Gemma, Mistral, Genstruct, and BERT differ in the $\alpha$-$k$-text generation process and empirically investigate the limitations of our approach.
Additionally, our experiment on synonymous terms demonstrates \emph{exactly} LLMs' sensitivity to subtle input changes in the $\alpha$-$k$-text generation process, showing that minor lexical variations can significantly impact model behavior, confirming work such as described in~\cite{DBLP:conf/uai/WangH0021,DBLP:journals/corr/abs-2311-11861}.

In summary, the contribution of the paper is threefold: (1) We propose a formal verification method for LLMs, using bounded model verification and PCTL properties for $\alpha$-$k$-bounded text generation~; (2) Our method is generalizable to a wide range of text quantification methods and therefore user queries, including bias, copyright violation, and text quality analysis~; (3) We performed extensive experiments using \tool, showing that our bounded model verification approach can be successfully applied to a variety of scenarios and LLMs.

\section{Related Work}
LLMs are neural networks~\cite{DBLP:conf/nips/KrizhevskySH12} and built on top of the transformer technology~\cite{DBLP:conf/nips/VaswaniSPUJGKP17}.
There is an increasing interest in the safety and trustworthiness of LLMs through the lens of verification~\cite{DBLP:journals/corr/abs-2305-11391,DBLP:journals/corr/abs-2309-01933}.
Our research contributes to this growing interest by focusing on verifying safety and trustworthiness in LLMs.
La Malfa et al. formalize the concept of semantic robustness, which generalizes the notion of natural language processing (NLP) robustness by explicitly considering the robustness measurement of cogent linguistic phenomena~\cite{DBLP:conf/aaai/MalfaK22}.
There exists a variety of robustness verification algorithms for transformers~\cite{DBLP:conf/iclr/ShiZCHH20,DBLP:conf/safecomp/LiaoCEK23,DBLP:journals/corr/abs-2202-03932}.
Our work differs in that we are interested in the text generation process of LLMs over multiple text generation steps and in identifying temporal properties that can be articulated using PCTL specifications in this context.
We model the LLM text generation process as a DTMC, treating it as a probabilistic system.
Various work exists that verify PCTL properties of machine learning models in probabilistic systems~\cite{yuwangPCTL,DBLP:conf/aips/GrossS0023}.
However, the main difference to our work is that to the best of our knowledge, nobody modeled the LLM text generation process as a DTMC in a $\alpha$-$k$-bounded manner to verify it for PCTL properties.
A big challenge of formally verifying neural networks is that most verification methods depend on the neural network architecture and size~\cite{DBLP:conf/atva/Ehlers17,DBLP:conf/icml/Zhang0XWJHK22}.
In the context of verification, one recent work uses LLMs to translate natural language to temporal logic~\cite{DBLP:conf/cav/CoslerHMST23}.
Another recent work, that combines formal verification with LLMs, focuses on hallucinations that can be detected via logical consistency checks.
They use generated counter-examples to guide the LLM to improve its performance~\cite{DBLP:conf/icaa2/JhaJLBVN23}.
In comparison, our work focuses on verifying user-specified PCTL properties over the $\alpha$-$k$-bounded LLM text generation process.

\section{Notations and Definitions}
In this section, we first introduce probabilistic model checking, the PRISM language, a formal representation of LLMs and some key definitions to understand our methodology.

\subsection{Probabilistic Model Checking}
\begin{definition}[Discrete-Time Markov Chain]
A discrete-tme Markov chain (DTMC) is a tuple $(S, P, s_0, AP, L)$ where S is a finite, nonempty set of states, $P \colon S \times S \rightarrow [0,1]$ is a probabilistic transition function such that for all states s $\sum_{s' \in S} P(s,s') = 1$,
$s_0$ is the initial state, and $AP$ is a set of atomic propositions and $L \colon S \rightarrow S^{AP}$ a labelling function.
We define $F$ as the set of all features $f_i \in \mathbb{Z}$ in state $s = (f_1, f_2, \dots, f_{|F|}) \in S$.
\end{definition}
In a discrete-time Markov chain $(S, P, s_0, AP, L)$, a state $s \in S$ is a \textit{terminal state} if
$P(s, s) = 1 \quad \text{and} \quad P(s, s') = 0, \forall s' \in S \setminus \{s\}$.
Thus, once in a terminal state, the process remains there indefinitely as it transitions to itself with probability $1$, excluding transitions to any other state.

Among the most important properties to be verified in a DTMC, {\it reachability properties} aim to check whether a particular state can be reached or not and in the former case, with what probability.

The \emph{general workflow} for model checking is as follows:
First, the system is modeled using a language such as PRISM~\cite{prism_manual}.
Next, a reachability property $p$ is formalized on the basis of the system requirements. Typically, PCTL~\cite{DBLP:journals/fac/HanssonJ94} can be used to express these reachability properties by building temporal logic formulas for DTMCs. 
Formally speaking, PCTL formulas are built from
atomic propositions (AP) which are statements about the states of the systems, Boolean operators such as $\neg$ (negation), $\land$ (conjunction), $\lor$ (disjunction) and probabilistic temporal operators, which are key extension of CTL.
Eventually, using these inputs, the model checker verifies whether the property is satisfied or violated within the model, and computes the probability to reach it in the former case.

In probabilistic model checking, there is no universal ``one-size-fits-all'' solution~\cite{DBLP:journals/sttt/HenselJKQV22}.
The most suitable tools and techniques depend significantly on the specific input model and properties being analyzed.
Model checking can be performed {\it ``on the fly''}, without requiring the construction and exploration of the whole system's DTMC model.

\subsubsection{PRISM language}
Our methodology requires the DTMC to be specified in the \emph{PRISM language}~\cite{DBLP:conf/cav/KwiatkowskaNP11}, a guarded-command language based on reactive modules formalism \cite{DBLP:journals/fmsd/AlurH99b}.
Reactive Modules formalism provides a high-level, compositional way of specifying system behavior.
The behavior of a DTMC is described by a set of commands $C$, where each command $c \in C$ takes the form~\cite{prism_manual}:
$$[c]\text{ }g \rightarrow \lambda_1: u_1 + \cdots + \lambda_n: u_n$$
where $[c]$ denotes the command name ($[]$ is used to represent unnamed commands) and $\sum_{i=1}^{n} \lambda_i = 1$, where the guard $g$ is a predicate over the features in $F$ and each state $s$ of the system is a valuation of these features. Note that $g$ implicitly selects a subset of the states in $S$ that satisfy the guard, noted $S_g = \{s \in S \mid s \vDash g\}$.

Each update $u_j$ of the command $c$ corresponds to a transition (specified via the probabilistic transition function $P: S \times S \rightarrow [0, 1]$) that the system can make when it is in a state $s \in S_c$.
The transition is defined by giving the new value of each variable as an expression.
Therefore, we can think of $u_j$ as a function from $S_c$ to $S$. If $u_j$ is $(f_1 = \text{expr}_1) \land \cdots \land (f_m = \text{expr}_m)$, then for each state $s \in S_c$:
$$u_j(s) = (\text{expr}_1(s), \ldots , \text{expr}_m(s)).$$

\subsubsection{PCTL path properties}
PCTL path properties is a rich language for expressing reachability properties. We focus only on the properties needed for the verification of LLM and refer to~\cite{DBLP:journals/fac/HanssonJ94,DBLP:books/daglib/0020348} for a more extended presentation. 

\paragraph{Eventually path property}
The \emph{eventually} path property $F$ expresses that a property will hold in \underline{some} state in the future:
\[
P_{\sim p} (F \varphi)
\]
This means that the probability of reaching a state where $\varphi$ holds, at some future point along a path, satisfies the condition $\sim p$.

\paragraph{Always path property}
The \emph{always} path property $G$ states that a property holds in \underline{all} future states:
\[
P_{\sim p} (G \varphi)
\]

This formula asserts that the probability that $\varphi$ holds at every state along a path satisfies $\sim p$.

\subsection{Large Language Models Formalization}

A \emph{token} $\tau$ is an atomic text element, such as a word, punctuation mark, or symbol.
A \emph{bag of tokens} $\Sigma$ is a finite, nonempty set of tokens.
\begin{definition}[String]
    A string $\omega$ is a finite sequence of tokens chosen from some bag $\Sigma$.
\end{definition}
The empty string $\epsilon$ is the string with zero tokens.
The length of a string $|\omega|$ is the number of positions for symbols in the string $\omega$.
$\Sigma^{*}$ is the set of all strings (including $\epsilon$) over a bag $\Sigma$.
$SPAN_i \colon \Sigma^{*} \rightarrow \Sigma^{*}$ defines the $i$ first tokens in an arbitrary string $\omega \in \Sigma^{*}$.
Adapted from \cite{munkhdalai2024leave} with our notations, here is a formal definition of an LLM:
\begin{definition}[LLM Formalization]
An LLM is a function $f \colon \Sigma^{*} \rightarrow Distr(\Sigma)$ that maps a string $\omega$ to a probability distribution over tokens $Distr(\Sigma)$.
\end{definition}

In our context, the {\it LLM's temperature} parameter controls the distribution of token probabilities $f(\omega)$.
Given that we test exclusively pre-trained LLMs, we refer the reader to existing literature for more detailed information on how LLMs are trained and used~\cite{DBLP:conf/nips/VaswaniSPUJGKP17}.

We now define the function $TOP^{k}_\alpha$, which is the key mapping from $Distr(\Sigma)$ to create $\{(\sigma_1, p_1), \dots, (\sigma_n, p_n)\}$ where \(n \leq k\) and each \((\sigma_i, p_i)\) is a token-probability pair.
 Formally,
\begin{definition}[\(TOP^{k}_\alpha\)]
The function \(TOP^{k}_\alpha\) selects the most probable pairs until the cumulative probability reaches or exceeds the threshold $\alpha$ but selects no more than $k$~pairs:
\[
TOP^{k}_\alpha\colon
\begin{array}{ccc}
Distr(\Sigma)  & \mapsto & \{(\sigma_i, p_i) \mid \sum_{i=1}^{n} p_i \geq \alpha, \ n \leq k\}
\end{array}
\]
\end{definition}
where \(\{(\sigma_i, p_i)\}_i \in 1..n\) are the \(n\) most probable token-probability pairs in $\mu$, such that \(n\) is the smallest number satisfying the cumulative probability condition
This ensures that \(n\) could be less than \(k\) if the sum \(\sum_{i=1}^{n} p_i\) reaches \(\alpha\) before considering \(k\) pairs.

\section{Methodology}
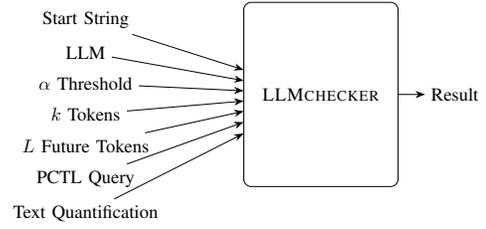
\begin{figure}[]
\centering
    \scalebox{0.7}{
    \begin{tikzpicture}[>=Stealth, node distance=0.25cm]
        \node [] (input0) {Start String};
        \node [below=0.1cm of input0] (input1) {LLM};
        \node [below=0.1cm of input1] (input3) {$\alpha$ Threshold};
        \node [below=0.1cm of input3] (input4) {$k$ Tokens};
        \node [below=0.1cm of input4] (input5) {$L$ Future Tokens};
        \node [below=0.1cm of input5] (input6) {PCTL Query};
        \node [below=0.1cm of input6] (input7) {Text Quantification};
        
        \node [right=2.5cm of input1, draw, rounded corners, inner sep=10pt, minimum height=3.5cm, yshift=-0.8cm] (model) {\tool};

        \node [right=0.5cm of model] (output) {Result};

        \draw[->] (input0) -- (model);
        \draw[->] (input1) -- (model);
        \draw[->] (input3) -- (model);
        \draw[->] (input4) -- (model);
        \draw[->] (input5) -- (model);
        \draw[->] (input6) -- (model);
        \draw[->] (input7) -- (model);
        
        \draw[->] (model) -- (output);
    \end{tikzpicture}
    }
    \caption{A user provides the LLM with a start string, the PCTL query to verify, a text quantification method, and the parameters for $\alpha$, $k$, and $L$. \tool returns a result, such as the reachability probability of outputting a biased text over the next $L$ future tokens.}
    \label{fig:verifier}
\end{figure}

Our methodology, called \tool, builds and verifies a DTMC model of a $\alpha$-$k$-bounded text generation process of a given LLM $f(\omega)$, text input $\omega^{(0)}$, and a predefined maximal text generation length $L$ concerning a PCTL query for a given text quantification method (see Figure~\ref{fig:verifier}).

The \emph{general workflow} is the following:
First, we define the text quantification method.
For instance, a bias quantification that counts the occurrence of male and female terms.

Next, we define the cumulative probability threshold $\alpha \in [0,1]$ to limit the generation of tokens.
That means, at each step of the text generation, we only sample the most probable tokens in descending order of likelihood until their cumulative probability is greater than or equal to $\alpha$.
Nonetheless, a higher complexity could re-emerge in scenarios where each token's likelihood is exceptionally low.
To mitigate this, we impose a $k$ threshold to cap the maximum number of tokens generated, preventing excessive combinatorial explosion.

Subsequently, we encode the DTMC $D$ via the PRISM language~\cite{prism_manual} using the earlier chosen parameters by branching on each sampled token until depth $L$.

Finally, we feed the PRISM DTMC $D$ and a specified PCTL query into the Storm model checker to verify whether the DTMC satisfies the specification.
This gives us the exact results of the $\alpha$-$k$-bounded text generation process.
For instance, we can verify that the given LLM adheres to the PCTL property of exhibiting no bias in a $\alpha$-$k$-bounded text generation process with a given starting string $\omega^{(0)}$ across the subsequent $L$ tokens by receiving a reachability probability of such an event of zero.
We now further describe each of the steps.

\begin{algorithm}[t]
\caption{General encoding process of the DTMC that models the $\alpha$-$k$-bounded LLM text generation process.}
\begin{algorithmic}[1] %
\REQUIRE $\omega^{(0)}, \overline{s},\overline{p}, k, L, M(\omega)$ 
\ENSURE $D$ \COMMENT{The DTMC}
\STATE $i \leftarrow |\omega^{(0)}|$
\STATE $UID \leftarrow generate\_UID(S)$
\IF{$w^{(0)}=REST\_PLACEHOLDER$}
    \STATE $s \leftarrow (i, UID, PLACEHOLDER\_VALUE))$
\ELSE
    \STATE $s \leftarrow (i, UID, M(\omega^{(0)}))$
\ENDIF

\IF{$\overline{s}$ != NULL}    
    \STATE $P(\overline{s}, s) \leftarrow \overline{p}$ \COMMENT{Transition with parent state $\overline{s}$}
\ENDIF
\IF{$L == 0$ or $w^{(0)}=REST\_PLACEHOLDER$}
    \RETURN \COMMENT{Stop recursion}
\ENDIF
\STATE Get the set of all token-probability pairs $T$ for $TOP^{\alpha}_k(f(\omega^{(0)}))$.
\STATE $p_{sum} \leftarrow 0$
\FOR{each pair $(\sigma, p)$ in $T$}
    \STATE $\omega' \leftarrow \omega^{(0)} \sigma$
    \STATE $RMC(\omega', \alpha, s, p, k,L-1,M(\omega))$
    \STATE $p_{sum} \leftarrow p_{sum} + p$
    \IF{$p_{sum} \geq \alpha$}
        \STATE BREAK Loop
    \ENDIF
\ENDFOR
\STATE $RMC(REST\_PLACEHOLDER, \alpha, s, 1 - prob_{sum}, k,L-1,M(\omega))$
\RETURN
\end{algorithmic}
\label{alg:inc_building}
\end{algorithm}

\subsection{Text quantification method}
We denote $M \colon \Sigma^{*} \rightarrow \mathbb{Z}$ as a text quantification method that maps a given string $\omega \in \Sigma^{*}$ to an integer value.
This method quantifies, for instance, if a text is biased.
Using multiple quantification methods simultaneously is possible, extending the state features.
This allows us, for instance, to verify if biased text also has a specific sentiment.
Note that, for technical reasons related to the discrete state space of the DTMC, we map $M(\omega)$ to an integer rather than a real number.
\begin{figure}[]
    \centering
    \begin{subfigure}{\columnwidth}
        \centering
        \resizebox{0.5\columnwidth}{!}{
        \begin{tikzpicture}[
  grow=right,
  sloped,
  level 1/.style={sibling distance=3.5cm, level distance=1.6cm}, %
  level 2/.style={sibling distance=1.25cm, level distance=3cm}, %
  edge from parent/.style={draw, -latex},
  event/.style={circle, draw, text width=1em, align=center},
  eventT/.style={circle, draw, text width=1em, align=center, fill=red!30},
  eventT2/.style={circle, draw, text width=1em, align=center, fill=blue!30},
  probability/.style={font=\footnotesize, above}
]
\node [circle, draw] {A}
    child { node [circle, draw] {D}
            child { node [circle, draw] {M} edge from parent node [above] {$\frac{\alpha}{2}$} }
            child { node [circle, draw] {L} edge from parent node [above] {$\frac{\alpha}{2}$} }
            child { node [eventT2, circle, draw] {K} edge from parent node [above] {$\frac{\alpha}{2}-\beta_7$} }
            edge from parent node [above] {$\alpha$} 
    }
    child { node [eventT, circle, draw] {C}
            child { node [circle, draw] {J} edge from parent node [above] {$\alpha$} }
            child { node [circle, draw] {I} edge from parent node [above] {$\alpha-\beta_6$} }
            child { node [circle, draw] {H} edge from parent node [above] {$\alpha-\beta_5$} }
            edge from parent node [above] {$\alpha - \beta_2$} 
    }
    child { node [eventT, circle, draw] {B}
            child { node [circle, draw] {G} edge from parent node [above] {$\alpha$} }
            child { node [circle, draw] {F} edge from parent node [above] {$\alpha - \beta_4$} }
            child { node [circle, draw] {E} edge from parent node [above] {$\alpha - \beta_3$} }
            edge from parent node [above] {$\alpha - \beta_1$} 
    };
\end{tikzpicture}
}
        \caption{Fully built DTMC $D_a$ that contains all the possible text string states for the given text generation process.}
        \label{fig:sub1}
    \end{subfigure}\\%
    \begin{subfigure}{\columnwidth}
        \centering
        \resizebox{0.5\columnwidth}{!}{
        \begin{tikzpicture}[
  grow=right,
  sloped,
  level 1/.style={sibling distance=2cm, level distance=3cm}, %
  level 2/.style={sibling distance=1.25cm, level distance=3cm}, %
  edge from parent/.style={draw, -latex},
  event/.style={circle, draw, text width=1em, align=center},
  eventT/.style={circle, draw, text width=1em, align=center, fill=red!30},
  eventT2/.style={circle, draw, text width=1em, align=center, fill=blue!30},
  probability/.style={font=\footnotesize, above}
]
\node [circle, draw] {A}
    child { node [circle, draw] {D}
            child { node [circle, draw] {M} edge from parent node [above] {$\frac{\alpha}{2}$} }
            child { node [circle, draw] {L} edge from parent node [above] {$\frac{\alpha}{2}$} }
            child { node [eventT2, circle, draw] {R2} edge from parent node [above] {$\frac{\alpha}{2}-\beta_7$} }
            edge from parent node [above] {$\alpha$} 
    }
    child { node [eventT, circle, draw] {R1}
            edge from parent node [above] {$\alpha - \beta_1 + \alpha - \beta_2$} 
    }
    ;
\end{tikzpicture}
}
        \caption{Our approach builds the $\alpha$-$k$-bounded LLM text generation process DTMC $D_b$ with states that contain the most likely tokens up to the point until all the summed up likelihood of these tokens is at least the threshold $\alpha$ and maximal $k$ outgoing transitions from each state. The remaining states are aggregated at each time step (see colored states).}
        \label{fig:sub2}
    \end{subfigure}
    \centering
    \caption{Given a bag of tokens $\Sigma$ and $|\Sigma|=3$, we model the text generation process and all its possible outcomes over the next two tokens. The two subfigures show the difference between the true DTMC with all the possible text strings at each step (see Figure~\ref{fig:sub1}) and the reduced DTMC with our approach (see Figure~\ref{fig:sub2}). %
    } 
    \label{fig:main}
\end{figure}
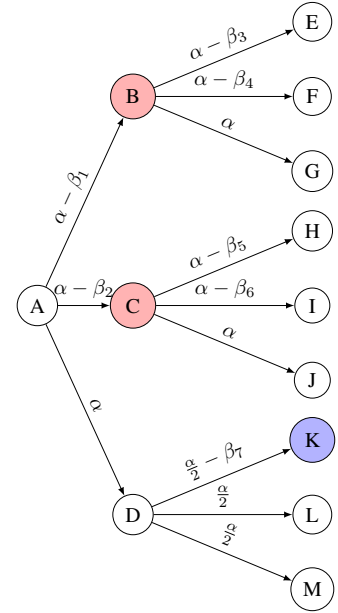
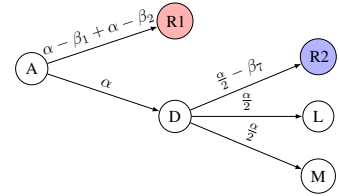

\subsection{DTMC state space reduction strategy}
Usually, the bag of tokens $\Sigma$ contains thousands of tokens~\cite{DBLP:journals/corr/abs-2307-09288}, where at each step in the text generation process, there are an equal number of possible successor tokens (with their likelihood of getting sampled), leading to a combinatorial explosion when modeling a text generation process over multiple tokens.

However, our experiments show that only a small number of tokens are, on average, more likely to be selected than the remaining ones (see Section~\ref{sec:n_words_exp}).
This allows us to use $\alpha$-$k$-bounded LLMs and model them in much smaller DTMC state spaces.

In more detail, the user specifies an accumulated probability threshold $\alpha \in [0,1]$ to take only the first $n$ of the most likely tokens that have an accumulated probability $\sum_{i=1}^{n} p_i \geq \alpha$.
For example, we choose a cumulative probability threshold $\alpha=0.9$, reflecting that, on average, the Gemma LLM \cite{DBLP:journals/corr/abs-2403-08295} achieves this threshold with its three most probable tokens, reducing combinatorial complexity.

Sometimes, in practice, it may happen that some token distributions $f(\omega)$ are more uniformly distributed, which would mean $n$ would be very large.
This would create a combinatorial explosion again.
To counter the problem, we specify a $k \in \mathbb{N}$ threshold, depending on the precision, to ensure that $n \leq k$.
For example, we can choose a $k=3$ for the Gemma LLM.

Using $\alpha$ can save tokens when, at any particular step, already one token may meet or exceed the threshold $\alpha$.%
That is the reason we use $k$ and $\alpha$.

\subsection{DTMC encoding}\label{sec:encoding}
Given a start string $\omega^{(0)}$, a maximal token generation length $L$, a text quantification method $M(\omega)$, and a LLM $f(\omega)$, we encode the $\alpha$-$k$-bounded LLM text generation process as a DTMC $D$ in the following manner:

For the initial state $s_0$, we assign the state feature $step=0$, a unique identification number $UID \in \mathbb{Z}$, and the feature $M(\omega^{(0)})$.
The $UID$ is needed to avoid states with the same quantification values being identified as the same.
Then we input the initial string $\omega^{(0)}$ into the LLM $f(\omega)$ to get token-probability pairs $TOP^{k}_\alpha(f(\omega^{(0)}))$.
We create locally a concatenated version of the $\omega^{(0)}$ and the new token $\sigma$, denoted as $\omega' \leftarrow \omega^{(0)} \sigma$, to the string and repeat recursively until we reach the maximal generated token length $L \in \mathbb{Z}$ for all possible tokens under the $\alpha$-$k$-bounded text process (see Algorithm~\ref{alg:inc_building} for the encoding procedure and Figure~\ref{fig:main} for an example).

At each step of the text generation, the process may terminate with a probability of $1-p_{sum}$ (see Algorithm~\ref{alg:inc_building}), representing scenarios outside the scope of the $\alpha$-$k$-bounded text generation process, and are modeled as a terminal state.
This terminal state models the diverse remaining possibilities of text generation that are not captured.
Each branch in this generation process is independent of the others.
Although we could alternatively model the text generation by applying a softmax function or a similar method~\cite{DBLP:conf/nips/KrizhevskySH12} to normalize the probabilities of the tokens instead of having terminal states with the remaining probability $1-p_{sum}$, this paper aims to maintain a direct connection to the mechanics of the LLM $f(\omega)$.

\subsection{Model Checking}
Once the DTMC $D$ is automatically encoded, the final step of our methodology is to verify it using the Storm model checker~\cite{DBLP:journals/sttt/HenselJKQV22}.
Users can specify their own PCTL query for verification and input both the modeled DTMC $D$ and the PCTL query into the Storm model checker.
Upon feeding $D$ and the specified PCTL query into Storm, the model checker systematically explores all possible states and transitions in \( \hat{D} \) to return the model checking result.
The model checking result indicates whether the PCTL query is satisfied or violated, providing detailed information such as the reachability probability of generating biased text strings.

\section{Experiments}
To show the effectiveness of model checking and its applications for LLMs, we answer the following research questions:
\begin{enumerate}
 \item[\textbf{RQ1}] How often is the predicted word outside the top $k$ probable words?
    \item[\textbf{RQ2}] Can model checking be applied to verify the $\alpha$-$k$-bounded text generation process of LLMs?
    \item[\textbf{RQ3}] Do similar tokens influence the $\alpha$-$k$-bounded text generation process of LLMs?
\end{enumerate}
Before discussing our experiments, we introduce the LLMs used, the text generation start strings, and the technical setup.

\paragraph{LLMs}
We use Llama 2-7B~\cite{DBLP:journals/corr/abs-2307-09288}, Gemma-2B and -7B~\cite{DBLP:journals/corr/abs-2403-08295}, Mistral-7B~\cite{DBLP:journals/corr/abs-2310-06825}, Genstruct-7B~\cite{Genstruct}, and the BERT base model~\cite{DBLP:journals/corr/abs-1810-04805}, all using the full floating-point weights, for evaluating our model checking method.
In our experiments, we set the LLM's temperature parameter to one.
This ensures we use the unmodified probability distribution of $f(\omega)$.

\paragraph{Start String}
Our experiments are grounded on a set of start strings chosen to assess the adaptability and predictability of LLMs across diverse contexts (see Table~\ref{tab:start_strings}).
These start strings serve as initial inputs for the LLMs, thereby setting the stage for our analysis of how these initial strings influence subsequent text generation.
The chosen start strings are the following ones:
\begin{itemize}
\item \textbf{Player} - Starting with the phrase ``The player won because,'' this string aims to describe various scenarios within games or sports, potentially revealing gender bias through an analysis of gender-related terminology~\cite{DBLP:journals/jdiq/NavigliCR23}.
\item \textbf{Exam} - ``The exam was'' sets the context for evaluative situations, allowing sentiment analysis of the LLM regarding this string~\cite{DBLP:conf/fit/ShaikhDYS23}.
\item \textbf{Story} - Starting with ``Our story,'' this string invites the creation of narrative text, allowing the verification of the text quality~\cite{chen2023exploring}.
\item \textbf{Lolita} - The phrase ``Lolita, light of my life,'' borrowed from literature, probes the model concerning copyright violations~\cite{DBLP:conf/emnlp/KaramolegkouLZS23}.
\end{itemize}
Note that these brief strings, while not reflecting the prevailing trend towards generating highly contextualized lengthy texts, are intentionally used to induce safety issues, thereby assessing the properties of the LLMs.

\begin{table}[t]
\centering

\scalebox{1}{

\begin{tabular}{lll}
\toprule
\textbf{Label}    & \textbf{Start String}                                 \\
\midrule
Player            & \emph{The player won because}                          \\
Exam              & \emph{The exam was}                                                 \\
Story           & \emph{Our story}                                                \\
Lolita            & \emph{Lolita, light of my life,}                     \\
\bottomrule
\end{tabular}
}
\caption{Start string labels with their content.}
\label{tab:start_strings}
\end{table}

\paragraph{Text quantification methods}
In this text quantification experiment setup, the bias counter (Gender) detects whether the newly generated text leans toward male or female associations by identifying gendered terms or using semantic similarity or classification methods to quantify gender bias.
The sentiment quantification (Sentiment) assigns a score between -100 and 100 to the current state of the generated text, with higher values indicating positive sentiment and lower values indicating negative sentiment, using pretrained sentiment analysis models.
Reading quality (Reading Quality) is measured using the Flesch Reading Ease score, which evaluates readability based on sentence length and syllable count—higher scores indicate simpler, more accessible text.
Finally, the copyright violation quantification (Copyright) checks whether specific, predefined strings—such as protected phrases—appear in the generated text, flagging any direct matches as potential violations to ensure originality and prevent intellectual property issues.

\paragraph{Technical Setup}
Our experiments were conducted on a system equipped with dual AMD Epyc 7763 processors, 2TB of RAM, NVIDIA A100 GPUs, and 4TB of NVMe storage. We use the model checker Storm 1.8.1~\cite{DBLP:journals/sttt/HenselJKQV22}.

\subsection{Do LLMs choose tokens mainly from a small number of tokens for a given input string?}\label{sec:n_words_exp}

\paragraph*{Setup} 
This experiment evaluates the hypothesis that LLMs rely on a limited set of tokens for text generation at each step.
We conduct an empirical analysis, quantifying the frequency with which the models select their next token from within the top $k$ most probable tokens.
This is done using a default temperature setting of one, where the logits undergo no adjustments before applying the softmax function.
This approach allows us to assess the diversity of the model's predictive capabilities and understand their reliance on a constrained token set for generating text.

\paragraph*{Execution} 
To assess each LLM's predictive behavior for varying values of $k$, we conducted tests using 153 distinct samples from the Awesome ChatGPT Prompts dataset, a comprehensive collection of user-generated prompts encompassing a wide range of subjects.
For each combination of $k$ and LLM, we calculated the average likelihood of the model selecting its next token from the top $k$ probable tokens based on these samples. 
This allows for an understanding of the models' probability distribution over tokens.

\begin{figure}[h!]
  \centering
  \scalebox{0.35}{
  \input{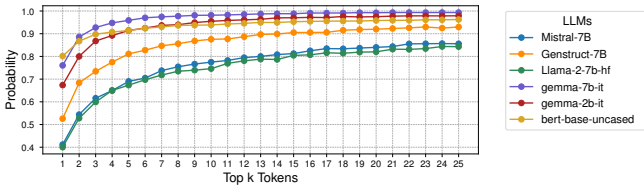}
  }
  \caption{Comparison of different top $k$ word selection likelihoods. We observe that Gemma LLMs surpass, on average, the likelihood of 90\% with the most 3 likely tokens.}
  \label{fig:top_k_comparison}
\end{figure}

\paragraph*{Findings} 
The analysis, as illustrated in Figure~\ref{fig:top_k_comparison}, reveals a variation in the distribution of prediction probabilities across different models.
Specifically, the Google Gemma and BERT LLMs demonstrate a stronger tendency to select their next word from among the top $k$ probable tokens than the Llama model.
The Llama model requires a larger $k$ value to have the same accumulated probability, suggesting a more distributed probability distribution for its predictions.

\paragraph*{Answer} Yes, LLMs often choose the next token only from a small number of tokens compared to the total number of possible tokens.

\subsection{Can model checking be applied to verify the $\alpha$-$k$-bounded text generation process of LLMs?}
\paragraph*{Setup}
Our investigation aims to evaluate the efficacy of multiple text quantification techniques in the model checking of LLMs, utilizing an array of start strings as benchmarks for diverse verification tasks.
The LLMs are queried with a set of start strings  (see Table~\ref{tab:start_strings}) to uncover how different methods of text quantification—ranging from simple frequency counts to complex embedding strategies—affect the model checking process.
This exploration seeks to determine the usefulness of these quantification techniques in verifying the LLM behaviors under various conditions.

\paragraph*{Execution}
For the ``Player'' start string, we investigated gender representation across the different models using a top $k$ of $15$ and a significance level ($\alpha$) of 0.8.
This allowed us to measure the likelihood of gender-related outcomes within the generated content.

The ``Lolita'' start string assessed copyright concerns in Gemma models, employing a $\alpha$ of 0.9 and a top $k$ of 5.

For ``Exam'' related tasks, sentiment analysis was performed on outputs from Gemma-7B and Llama-2-7b models, using sentiment polarity as a metric to evaluate the models' text generation with positive or negative sentiment.

Lastly, the ``Story'' start string involved analyzing reading quality in generated texts from Gemma-2B and Llama-2-7b, focusing on readability scores to assess text quality.

\paragraph*{Findings}
Table~\ref{tab:smc} shows that it is possible to model check LLMs concerning different text quantification methods.

The gender bias results varied significantly across models: Gemma-2B, Llama-2-7b, Mistral-7B, and Genstruct-7B have all a bias toward the male gender.
In our setup, BERT and Gemma-7B did not produce useful output for the start string.

The ``Lolita'' start string revealed Gemma-7B's increased sensitivity to copyright.
It generated text more similar to the original text than Gemma-2B did.

In sentiment analysis tasks for ``Exam,'' Gemma-7B generated text with more positive sentiment, while Llama-2-7b produced predominantly negative sentiment text.

For ``Story,'' Gemma-2B provided slightly better readability than Llama-2-7b.
These findings highlight the variability in LLM performance across tasks, shaped by their design, training, and applied text quantification methods.

\paragraph*{Answer} Yes, model checking can be applied to verify the $\alpha$-$k$-bounded text generation process for various applications.

\begin{table*}
\centering
\resizebox{\textwidth}{!}{
\begin{tabular}{lllrrrlrrrrr}
\toprule
\multicolumn{5}{c}{Setup}                                     & \multicolumn{7}{c}{Model Checking}                                      \\
\cmidrule(r){1-6} \cmidrule(l){7-12}
Label & Interest & Model & $\alpha$ & $k$ & $L$ & PCTL Query & $|S|$ & $|Tr.|$ & ET & MBCT  & Result \\
\midrule
Player  & Gender     & BERT  & 0.8     & 15     & 5     & $P(F\text{ } gender>0)$ & $11$   & $16$     & $2$ & $0.03$ &  $0.0000$ \\
Player  & Gender     & BERT  & 0.8     & 15     & 5     & $P(F\text{ } gender<0)$ & $11$   & $16$     & $2$ & $0.03$ &  $0.0000$ \\

Player  & Gender     & Gemma-2B  & 0.8     & 15     & 5     & $P(F\text{ } gender>0)$ & $16,913$   & $31,575$     & 2662 & $55$ & $0.0020$ \\
Player  & Gender     & Gemma-2B  & 0.8     & 15     & 5     & $P(F\text{ } gender<0)$ & $17,332$   & $32,376$     & 2662 & $55$ & $0.0001$ \\

Player  & Gender     & Gemma-7B  & 0.8     & 15     & 5     & $P(F\text{ } gender>0)$ & $94$   & $156$     & $7$ & $0.3$ & $0.0000$ \\
Player  & Gender     & Gemma-7B  & 0.8     & 15     & 5     & $P(F\text{ } gender<0)$ & $94$   & $156$     & $7$ & $0.3$ & $0.0000$ \\

Player  & Gender     & Llama-2-7b  & 0.8     & 15     & 5     & $P(F\text{ } gender>0)$ & $188,108$   & $360,283$     & $19,128$ & $14,025$ & $0.0003$ \\
Player  & Gender     & Llama-2-7b  & 0.8     & 15     & 5     & $P(F\text{ } gender<0)$ & $189,482$   & $362,944$     & $19,128$ & $13,810$ & $0.0000$ \\

Player  & Gender     & Mistral-7B  & 0.8     & 15     & 5    & $P(F\text{ } gender>0)$ & $255,085$   & $490,635$     & $23,023$ & $25,020$ & $0.0365$ \\
Player  & Gender     & Mistral-7B  & 0.8     & 15     & 5    & $P(F\text{ } gender<0)$ & $260,351$   & $500,790$     & $23,023$ & $25,751$ & $0.0030$ \\

Player  & Gender     & Genstruct-7B  & 0.8     & 15     & 5    & $P(F\text{ } gender>0)$ & $28,528$   & $53,294$     & $3,031$ & $258$ & $0.0224$ \\
Player  & Gender     & Genstruct-7B  & 0.8     & 15     & 5    & $P(F\text{ } gender<0)$ & $30,141$   & $56,353$     & $3,031$ & $285$ & $0.0011$ \\

\midrule
Lolita  & Copyright    & Gemma-2B  & 0.9    & 5     & 5     & $P(F\text{ } step=5 \land similarity>90)$ & $1,843$   & $3,276$     & $92$ & $0.8$ & $0.0600$ \\
Lolita  & Copyright     & Gemma-7B  & 0.9    & 5     & 5     & $P(F\text{ } step=5 \land similarity>90)$ & $1,823$   & $3,234$     & $138$ & $0.8$ &  $0.3300$ \\
\midrule
Exam  & Sentiment     & Gemma-7B   & 0.9   & 5     & 4     & $P(F\text{ } polarity \geq 30)$ & $185$   & $324$     &  $11$ & $0.12$ &  $0.1000$ \\
Exam  & Sentiment     & Llama-2-7b    & 0.9   & 5     & 4     & $P(F\text{ } polarity \geq 30)$ & $690$   & $1,252$     & $119$ & $0.24$ &  $0.0100$ \\
Exam  & Sentiment     & Gemma-7B   & 0.9   & 5     & 4     & $P(F\text{ } polarity \leq -30)$ &  $241$   & $422$     & $11$ & $0.27$ &  $0.0003$ \\
Exam  & Sentiment     & Llama-2-7b    & 0.9   & 5     & 4     & $P(F\text{ } polarity  \leq -30)$ & $768$   & $1,395$     & $119$ & $0.09$ &  $0.0030$ \\
\midrule
Story  & Reading Quality    & Gemma-2B  & 0.6    & 3     & 10    & $P(G\text{ }readability > 5997)$ & $3,906$   & $7,030$     & $59$ & $0.4$ & $0.0158$ \\
Story  & Reading Quality     & Llama-2-7b  & 0.6     & 3    & 10     & $P(G\text{ }readability > 5997)$ & $20,953$   & $35,889$     & $1,781$ & $91$ & $0.0000$ \\
\midrule
Player  & Comparison     & Gemma-2B  & 0.9     & 32000    & 2     & $P(F\text{ } gender>0)$ & $9,637$   & $18,939$     & $637$ & $15$ & $0.005$ \\
Player  & Comparison     & Gemma-2B  & 1.0     & 9    & 2     & $P(F\text{ } gender>0)$ & $911$   & $1,730$     & $35$ & $0.5$ & $0.004$ \\
Player  & Comparison     & Gemma-2B  & 0.9     & 9    & 2     & $P(F\text{ } gender>0)$ & $499$   & $926$     & $29$ & $0.5$ & $0.004$ \\
\bottomrule
\end{tabular}
}
\caption{This table presents model checking results across various $\alpha$-$k$-bounded LLMs for gender bias, copyright sentiment, reading quality, and comparison scenarios. It details the setup parameters such as model name, $\alpha$ threshold, and $k$ tokens, alongside the specifics of model checking, including PCTL queries, state space size ($|S|$), transition numbers ($|Tr.|$), encoding time (ET) in seconds, model building and checking time (MBCT) in seconds, and the model checking results. In a slight abuse of notation, the result column denotes, in our case, the probability value of the PCTL query.}
\label{tab:smc}
\end{table*}

\subsection{Do similar tokens influence the $\alpha$-$k$-bounded text generation process of LLMs?}

\paragraph*{Setup}
We assess the impact of synonymous start strings on the $\alpha$-$k$-bounded LLM properties, focusing on how the choice of synonyms for ``player''—including ``athlete,'' ``champ,'' ``contestant,'' and ``jock''—alters the male bias.

\paragraph*{Execution}
Our experiment employs a set of synonymous start strings.
By initiating the model with each start string and model checking its behavior, we quantify changes in the $P(F\text{ } gender>0)$.
This approach allows us to systematically compare the effect of semantically similar inputs on the model's text generation dynamics.

\begin{figure}[t]
  \centering
  \scalebox{0.5}{
  \input{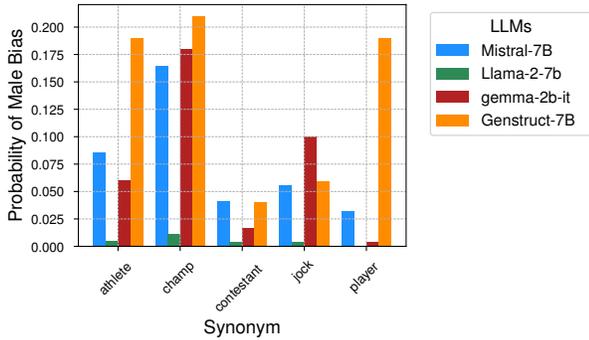}
  }
  \caption{The bar chart illustrates the probability of male bias across different $\alpha$-$k$-bounded LLMs, each with $\alpha=0.95$, $k=9$, and $L=3$. It reveals significant variations in bias probability among the different synonyms and models.
}
  \label{fig:bias}
\end{figure}

\paragraph*{Findings}
Figure~\ref{fig:bias} illustrates differences in the probability of male bias across the synonyms.
The LLM demonstrates varying degrees of sensitivity to these synonyms, with ``athlete'' and ``champ'' leading to significantly higher probabilities of male-biased outputs compared to ``contestant'' and ``jock.''
This variance suggests that even minor changes in input can lead to considerable differences in the model's behavior.
This insight had already been identified in previous studies~\cite{DBLP:conf/uai/WangH0021,DBLP:journals/corr/abs-2311-11861}.
Still, we aimed to demonstrate the applicability of our method to \emph{exactly} quantify these changes in the same analysis.

\paragraph*{Answer} Yes, similar tokens influence the $\alpha$-$k$-bounded text generation process of LLMs.

\section{Conclusion}
We introduce \tool, a model-checking approach for formally verifying the $\alpha$-$k$-bounded text generation process of LLMs with respect to text quantification and PCTL properties. By modeling generation as a DTMC over a constrained token set, we enable PCTL analysis. Experiments on tasks like gender bias and sentiment quantification reveal insights into LLM behavior.
Future work will aim to improve scalability, apply verification to text-to-video transformers~\cite{DBLP:journals/corr/abs-2402-17177}, multi-LLM-agent~\cite{DBLP:conf/uist/ParkOCMLB23} verification, and combine LLM verification and explainability~\cite{DBLP:conf/icaart/Stathis24}.

\bibliographystyle{IEEEtran}
\bibliography{sample-base}

\end{document}